\def\year{2018}\relax
\documentclass[letterpaper]{article} 
\usepackage{aaai18} 
\usepackage{times} 
\usepackage{helvet} 
\usepackage{courier} 
\usepackage{url} 
\usepackage{epsfig}
\usepackage{graphicx} 
\usepackage{caption}
\usepackage{amsmath}
\usepackage{amsfonts}
\usepackage{amssymb}
\usepackage{bbm}

\begin{document}
%
 \title{A Categorisation of Post-hoc Explanations for Predictive
 Models} \author{
 John Mitros and Brian Mac Namee\\
 \{ioannis, brian.macnamee\}@insight-centre.org\\
 School of Computer Science\\
 University College Dublin, Dublin, IR\\
 }
\maketitle

\begin{abstract}
  The ubiquity of machine learning based predictive models in modern
  society naturally leads people to ask how trustworthy those models
  are?  In predictive modeling, it is quite common to induce a
  trade-off between accuracy and interpretability.  For instance,
  doctors would like to know how effective some treatment will be for
  a patient or why the model suggested a particular medication for a
  patient exhibiting those symptoms? We acknowledge that the necessity
  for interpretability is a consequence of an incomplete formalisation
  of the problem, or more precisely of multiple meanings adhered to a
  particular concept.  For certain problems, it is not enough to get
  the answer (what), the model also has to provide an explanation of
  how it came to that conclusion (why), because a correct prediction,
  only partially solves the original problem. In this article we
  extend existing categorisation of techniques to aid model
  interpretability and test this categorisation.
\end{abstract}


\section{Introduction}
Due to the technological advancements across all aspects of modern
society many of our routine daily decisions are now either delegated
to, or driven by, algorithms. These algorithms, for example, decide
which emails are considered spam, if our credit loan application is
going to be approved, and whether our commute between locations
increases the probability of avoiding a traffic jam or an
accident. These ubiquitous algorithms, however, not only provide
answers but also raise questions that our society needs to
address. For instance, if an algorithm denies an applicant credit,
that applicant would most certainly want to be informed about why that
decision made to understand the adjustments they should make in order
to receive a better outcome for their next application, or to
challenge the decision. Similarly, a patient receiving medical
treatment would appreciate understanding the evidence for a diagnosis,
by what degree the diagnostic process is automated, and how much the
clinician trusts the algorithmic process.

The algorithms we are focused on in this article are supervised
machine learning algorithms that are used to build predictive models
\cite{kelleher2015}. Fundamentally, supervised machine
learning algorithms learn to distinguish patterns in input space by
associating inputs, $\vec{x}$, to outputs, $\vec{y}$. During the
process of training a model, an algorithm strives to optimize a
performance criterion using example data or past experience (for a
more formal definition we point the reader
to~\cite{mitchell1997}). Since predictive models have already
penetrated critical areas of society such as healthcare, justice
systems, and the financial industry, it is necessary to understand how
they arrive at decisions, to verify the integrity of the decision
making process, and to ensure that processes are in accordance with
the ethical and legal requirements. More succinctly, models need to be
\emph{interpretable} but this requirement poses a series of recent
challenges and trends regarding their design and
implementation details~\cite{holzinger2018}.

As we have already mentioned, interpretability is an important factor
for predictive models and the broader machine learning (ML) community,
but its importance derives from the scenarios in which machine
learning models are applied. For instance, it is crucial to have
interpretable explanations from predictive models assisting with
diagnosis in the medical domain to provide additional information to
domain experts (doctors) and guide their decisions, but one could
argue that the importance of interpretability fades when considering
customers being recommended products while browsing online shops.

\section{Defining Interpretability}
\begin{figure}[h]
  \centering \includegraphics[width=\columnwidth]{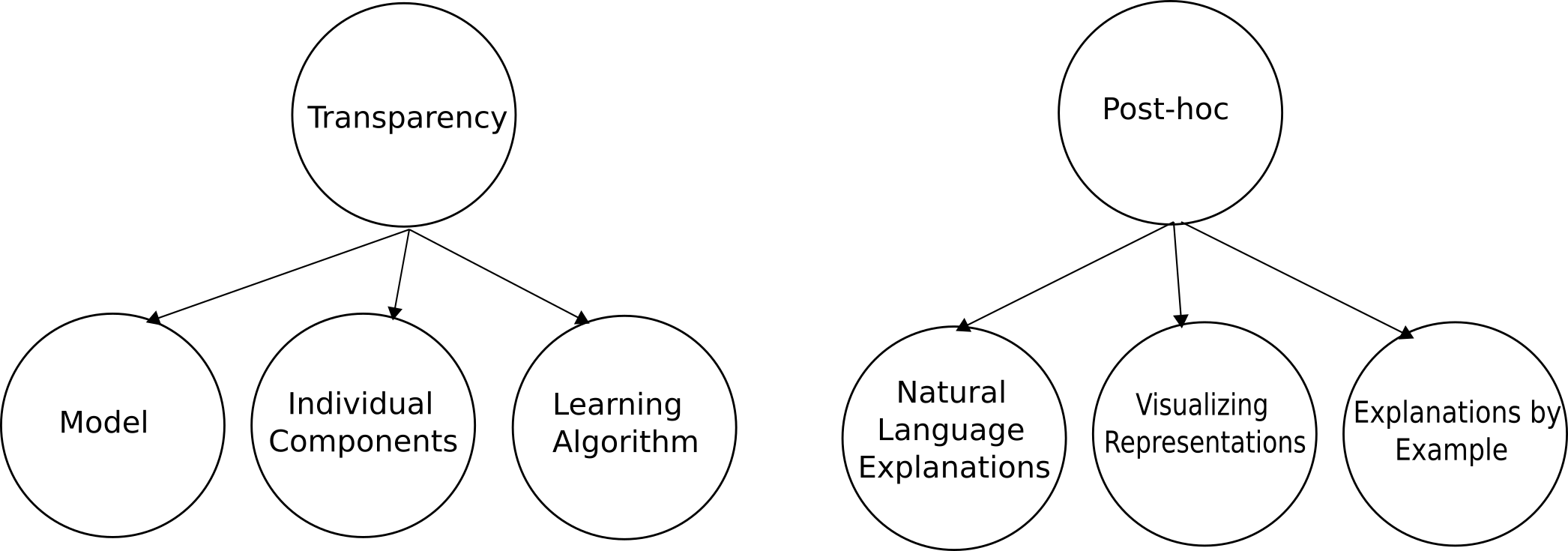}
  \caption{The scope of interpretability defined by
    Lipton~\cite{lipton2016}}
  ~\label{fig:categories}
\end{figure}

To discuss interpretability, first we need to define it. Throughout
the rest of the article we will use the definition by
Miller~\cite{miller2017}: ``\textit{interpretability is the degree to
  which a human can understand the cause of a decision}''.
Lipton~\cite{lipton2016} goes
further and divides the scope of interpretability into two categories:
\emph{transparency} and \emph{post-hoc explanations}. Transparency
refers to the intrinsic underlying mechanism of a model's inner
workings. According to
Lipton~\cite{lipton2016} this
could potentially describe mechanisms at the level of the model,
component, or algorithm. On the other hand post-hoc explanations aim
to provide further information about the model by uncovering the
importance of its parameters. A schematic representation of this
categorization is shown in Figure~\ref{fig:categories}.

\begin{figure}[h]
  \centering \includegraphics[width=0.7\columnwidth]{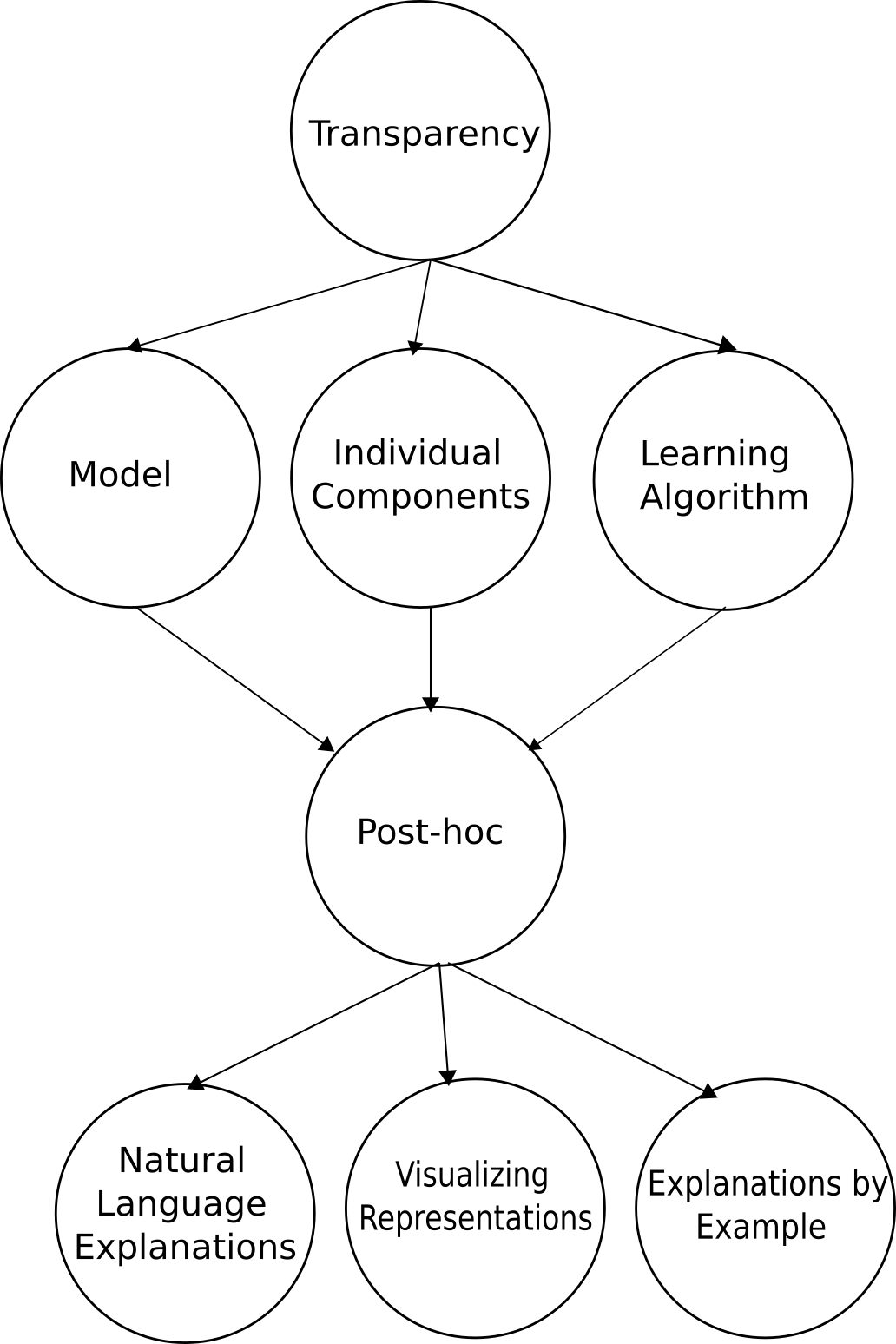}
  \caption{A hierarchical view of the scope of interpretability.}
  ~\label{fig:hierarchy}
\end{figure}

Although it might not be evident at first sight we can visualize the
two categories as being interconnected instead of separate, as shown
in Figure~\ref{fig:hierarchy}. This is possible due to the large
variety of post-hoc methods that are available and that can be
integrated at different levels of the life-cycle of a model.

\begin{figure}[h]
  \centering
  \includegraphics[width=\columnwidth]{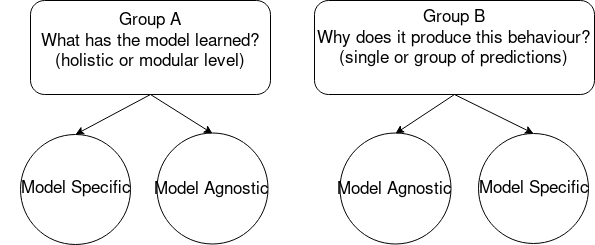}
  \caption{Grouping of post-hoc interpretable methodologies.}
  ~\label{fig:posthoc_groups}
\end{figure}

In this article we are concerned with the post-hoc explanation
category of interpretability. This category can be further subdivided
into their own groups. For the purpose of this study we will call them
\emph{Group A} and \emph{Group B}. Approaches in Group A address the
question of what has the model learned, either on a holistic or
modular level. Approaches in Group B are concerned with why the model
produced a specific behaviour, either for a single prediction or a
group of predictions. Each of them can be further subdivided into two
additional subgroups: \emph{model specific} that are tightly
intertwined with a specific modelling algorithm and \emph{model
  agnostic} approaches that can be applied to models trained using any
algorithm. A schematic of these subdivisions is presented in
Figure~\ref{fig:posthoc_groups}.

The remainder of this article tests this categorisation by describing
key approaches in Group A and Group B, as just described. We briefly
describe these approaches, draw connections and distinctions between
them, and identify promising areas for future work.

\section{Group A:~What has the model learned?}
~\label{sec:GroupA} Approaches that we categorise under Group A,
provide additional information about a trained predictive model,
either holistically or at a modular level. We first review model
agnostic approaches and then focus on model specific ones.

\subsection{Model Agnostic}
~\label{sec:GroupA_agnostic} Model agnostic approaches from Group A
can provide overall interpretations of any black box model. These
interpretations could be in regards to the overall model or to its
individual components. The most common type of approach in this
category is \emph{rule induction}, in which interpretable rules are
extracted from a trained black box model to describe its important
characteristics. Yang et al
~\cite{yang2018}, for example, introduced
the use of compact binary trees to represent important decision rules
that are implicitly contained within black box models.  To retrieve
the binary tree of explanations the authors recursively partition the
input variable space maximizing the difference in contribution of
input variables averaged from local explanations between the divided
spaces to retrieve the most important rules that represent the
magnitude of changes in the output variable due to one unit of change
in the input variable. This is also known as \emph{sensitivity
  analysis} in fields such as physics or biology.

The \emph{Black Box Explanation through Transparent Approximations}
(BETA)~\cite{lakkaraju2017},
approach is another rule induction method, closely connected to
earlier work presented
by~\cite{lakkaraju2016}. BETA learns a
compact two-level decision set in which each rule explains part of the
model behaviour unambiguously. BETA includes a novel objective
function so that the learning process is optimized for high fidelity
(increased agreement between explanations and the model), low
unambiguity (less overlaps between decision rules in the explanation),
and high interpretability (the explainable decision set is lightweight
and small). These aspects are combined into one objective function
that is optimized.

The approaches described so far can be applied to predictive models,
but there are also rule induction approaches for more complex models.
Penkox~\cite{penkov2017}, for example,
introduced the $\pi$-machine, which can extract LISP-like programs
from observed data traces to explain the behaviour of agents in
dynamical systems whose behaviour is driven by deep Q network (DQN)
models trained using reinforcement learning. The proposed method
utilizes an optimisation procedure based on backpropagation, gradient
descent, and $A^{\ast}$ search in order to learn programs and explain
various phenomena.

\subsection{Model Specific}
~\label{sec:GroupA_specific} Model specific methods from Group A are
designed specifically to derive overall interpretations of what models
trained using specific algorithms, or families thereof, have
learned. One compelling example is \emph{maximum variance total
  variation denoising} (MVTV)
~\cite{tansey2017} which is a
regression approach that explicitly gives interpretability of the
model which is of high priority and is useful when simple linear or
additive models fail to provide adequate performance. Recently there
has been an additional effort to extend a class of algorithms with
high classification accuracy rate such as Deep Neural Networks (DNN)
in order to extract insightful explanations in the form of decision
diagram rules. An example of such an effort is described
in~\emph{learning customized and optimized lists of rules with
  mathematical
  programming}~\cite{rudin2018},
providing a formal approach in order to derive interpretable lists of
rules. The methodology empowers solving a custom objective function
along with user defined constrains utilizing mixed integer programming
(MIP). This provides two immediate benefits, fist, the pruning of the
large space of the derived rules and second, the guaranteed of
selecting the most optimal rules that closely describe the
classifier's inner process.

Another approach which strives to extract interpretations from DNN is
presented in~\emph{this looks looks like that: deep learning for
  interpretable image
  recognition}~\cite{chen2018}. Here the emphasis
is based on defining a reasoning process similar to humans when
deciding if two objects belong to the same group based on common
characteristics exhibited among them. Thus, the methodology is trying
to discover a combination of exemplar patches from an image along with
its prototypes during the classification process. This ensures that
the final classification process adheres to principles closely related
to human decision making, therefore the final classification process
is interpretable by design. One such a example is showing and image of
a cat to a human and asking them to describe the process by which they
came to the final conclusion. Most probably they would focus on
particular areas of the image while making the final decision based on
previous experience, and excluding non similar objects. In
addition,~\emph{bayesian patchworks: an approach to case-based
  reasoning}~\cite{moghaddass2018} can be thought of as an
extension of the previous methodology by designing models that can
mimic logical processes resembled in individuals while reasoning about
previous examples. The main idea is based on deriving a generative
hierarchical model who's final classification decision is based on
examining similar examples exhibiting common features to identify a
common ancestor. Going back to our earlier example of identifying an
image of a cat, that would translate into a process of, (1) finding
previous examples with common feature vectors (i.e.~its neighbours),
(2) deciding how many of the neighbours share important properties,
(3) identifying those neighbours that are connected in terms of their
features (i.e.~share a common ancestor), (4) assign a label to the new
example based on the influence of its neighbours and ancestor
(e.g. similar to a voting approach but from a hierarchical
perspective).

\subsection{Summary}
~\label{sec:GroupA_conclusion} In summary one can interpret Group A as
being the focus of investigating interpretable techniques providing
either a holistic or modular view of the behaviour of a black-box
model. The main idea underlying these efforts is to provide a coherent
narrative through the process of a logical sequence which the end user
can interpret as part of the overall process. Thus, approaches in this
direction quite often utilize methods that either can provide this
narrative such as decision rules, decision sets or inherently
transparent methods such as regression variants. Similar efforts can
be found in the works
of~\citeauthor{lakkaraju2016}\;\shortcite{lakkaraju2016},~\citeauthor{condry2016}\;\shortcite{condry2016}
and~\citeauthor{letham2015}\;\shortcite{letham2015}.

\section{Group B:~Why does the model produce this behaviour?}
~\label{sec:GroupB} In contrast to those in Group A that seek to
explain a global view of what a model has learned, approaches in Group
B seek to explain the reasons behind a specific prediction for a
specific test instance. Again we summarise both model agnostic and
model specific methods.

\subsection{Model Agnostic}
~\label{sec:GroupB_agnostic}

The \emph{local interpretable model agnostic explanations} (LIME)
~\cite{ribeiro2016} approach is a seminal
example of a model agnostic Group B method. LIME is based on two
simple ideas: perturbation and locally linear approximation. By
repeatedly perturbing the values of the input $\vec{x}$ and observing
the effect this has on the output $\vec{y}$ an understanding can be
developed of how different inputs relate to the original output. LIME
performs a perturbation of a specific input and then trains a linear
model to describe the relationships between the (perturbed) inputs
$\acute{\vec{x}}$ and the predicted outputs $\hat{\vec{y}}$. The
simple linear model approximates the more complex original black box
model locally in the vicinity of the prediction that is to be
explained. The linear model is used as an explanation for the original
prediction.

Approaches similar to LIME are also presented in
~\cite{baehrens2010} and~\cite{robnik-sikonja2008}. Anchors~\cite{ribeiro2018}
is a particularly interesting extension to LIME that adds rule
induction to make the explanations more compelling. The ASTRID
method~\cite{henelius2017} provides
a form of post-hoc interpretability to gain insight into how a model
reaches specific predictions by examining attribute interactions in
the dataset.

\subsection{Model Specific}
~\label{sec:GroupB_specific} It can be more fruitful to build
approaches that only apply to specific model types as this allows the
inherent structure and learning mechanisms of these models to be
utilised in generating explanations. This section describes
interesting examples taking this approach.

The recent resurgence in interest in deep neural network approaches
\cite{goodfellow2016} has seen parallel interest in developing
explanation approaches for these deep network models. Montavon et al
\cite{montavon2017}, for example, propose a
novel methodology for interpreting generic multilayer neural networks
by decomposing the network classification into contributions of its
input elements. Their methodology is applicable to a broad set of
input data, learning tasks and network architectures. Explanations are
generated as heatmaps overlaid on input artefacts (such as images)
that indicate the areas of those artefacts that contributed most
strongly to the model output. LSTMVis~\cite{strobelt2016}, is a
visual analysis tool for recurrent neural networks with a focus on
understanding hidden dynamic states learned by these models. LSTMVis
also relies heavily on the use of visualisations to communicate
explanations.

Ensemble methods can also be difficult to interpret and specific
approaches for these types of models have been developed.  SHAP
(SHapely Additive
exPlanation)~\cite{lundberg2018},
for example, addresses a fundamental problem with feature attribution
when interpreting ensemble predictive models such as gradient boosting
and random forests, due to the fact that they are inconsistent and can
lower a feature's assigned importance. SHAP uses approaches form game
theory to recognise genuinely important features.

\subsection{Summary}
~\label{sec:GroupB_conclusion} Overall the aim of Group B, in contrast
to Group A, is to provide explanations regarding the behaviour of a
black-box model based on the predictions emitted by the model (single
or multiple predictions). The main idea is based on examining features
or interactions thereof in order to understand which pose the highest
shift in regards to the black-box model's behaviour. This approach is
known as sensitivity analysis in fields such as physics and biology
and adheres to the study of uncertainty in the output of a
mathematical model or system related to different sources of
uncertainty in its inputs. Similar approaches can be found in
~\cite{chen2018},~\cite{ancona2017}~\cite{kahng2017}~\cite{samek2017}~\cite{qi2017}~\cite{liu2017}~\cite{lapuschkin2016}~\cite{zhou2016}~\cite{mahendran2016}
and~\cite{li2016}.

\section{Conclusion}
~\label{sec:conclusion} In summary in this article we are trying to
provide an overview of the multifaceted aspects of
interpretability. Overall we proposed a hierarchical view of
interpretability adopted
from~\cite{lipton2016} with minor
modifications. Despite the fact that the emphasis has been on
introducing and disseminating post-hoc methodologies we have always
strove for a balance either by pointing the reader to the equivalent
bibliography or by introducing methodologies that could potentially
fall under the transparency group as well.

A recurring theme we have noticed throughout the existing literature is
the lack of clear distinction between the notions of interpretability
and debug(ability). It seems that most of the existing methodologies
provide a way of debugging a predictive model and implicitly deriving
the necessary interpretations. This constitutes an additional
confusion to the reader and overall for the community as well.

Finally, we would like to point the reader to an interesting direction
of research on constructing bug free predictive
models,~\cite{selsam2017}, falling under the
group of transparency methodologies, which potentially holds a promise
on possibly alleviating the need for interpretability. Furthermore,
work towards an axiomatic treatment of
interpretability~\cite{bodenhofer2000} has
already paved the path in which the community should be invested, in
order to alleviate the multifaceted properties of interpretability
exhibited until recently. We hope that this article can serve as a
guide in navigating the different aspects of interpretability to the
interested reader.



 \section{Acknowledgments}
 This work has been supported by a research grant by Science
 Foundation Ireland under grant number SFI/15/CDA/3520. We
 would also like to thank Nvidia for their generours donation of
 GPU hardware in support of our research.



\bibliographystyle{aaai}
\bibliography{references}
\end{document}